\documentclass{article}
\usepackage{authblk}
\usepackage{amsmath}
\usepackage{amsfonts}
\usepackage{dsfont}  
\usepackage{pdfpages}
\usepackage{pgf}

\usepackage[letterpaper]{geometry}
\usepackage{natbib}
\usepackage{graphicx}
\usepackage{multirow}
\usepackage{amsmath, amsfonts, amssymb}
\usepackage{amsthm}
\usepackage{amscd}
\usepackage{times}
\usepackage{booktabs}
\usepackage{url}

\newtheorem{definition}{Definition} 
\newtheorem{lemma}{Lemma} 

\usepackage[doublespacing]{setspace}

\title{Matrix sketching for supervised classification with imbalanced classes}
\author[1]{Roberta Falcone\footnote{mailto: \texttt{roberta.falcone3@unibo.it}}}
\author[1]{Angela Montanari}
\author[1]{Laura Anderlucci}
\affil[1]{Department of Statistical Sciences, University of Bologna, Italy.}

\date{}

\begin{document}
\maketitle

\begin{abstract} 
Matrix sketching is a recently developed data compression technique.
An input matrix $A$ is efficiently approximated with a smaller
matrix $B$, so that $B$ preserves most of the properties of $A$ up to some
guaranteed approximation ratio. In so doing numerical operations on
big data sets become faster. Sketching algorithms generally use random
projections to compress the original dataset and this stochastic generation
process makes them amenable to statistical analysis. The statistical properties of sketching algorithms have been widely studied in the context of multiple linear regression. In this paper we propose matrix sketching as a tool for rebalancing class sizes in supervised classification with imbalanced classes. It is well-known in fact that class imbalance may lead to poor classification performances especially as far as the minority class is concerned.
\end{abstract}


\section{Introduction}
	In many practical contexts, observations have to be classified into two classes of remarkably distinct size. \\
	In such cases, many established classifiers often trivially classify instances into the majority class achieving an optimal overall misclassification error rate. \\	
	This leads to poor performance
	in classifying the minority class, the correct identification of which is usually of more practical interest.

The presence of imbalanced classes in the big data context also poses relevant computational issues. If the dataset contains thousands or millions of observations from the majority class for each example from the minority one, many of the majority class observations are redundant. Their presence increases the computational cost with no advantage in terms of classification accuracy.\\
The problem of imbalanced classes is very common in modern classification problems and has received a great attention in the machine learning literature \citep{chawla2004special}.

	The error rate (or its complement, the accuracy) is the most widely used measure of classifier performance. However, it inevitably favors the majority class when the misclassification error has the same importance for the two classes. On the contrary, when the error in the minority class is more important than the one of the majority class, the receiver operating characteristic (ROC) curve and the corresponding area under the curve (AUC) are commonly suggested. \\
	The ROC curve is a plot of the true positive rate (\textit{sensitivity}) versus the false positive rate ($\mathit{1-}$\textit{specificity}) and hence a higher AUC generally indicates a better classifier.  The ROC is obtained by varying the discriminant threshold, while the error rate is obtained at an optimal discriminant threshold. Therefore, AUC is independent of the discriminant threshold, while the accuracy is not.

The literature on the problem of supervised classification is very broad and methodological solutions follow two main streams. People either suggest to modify the loss function used in the construction of the classification rule or propose to re-balance the data.

The first solution requires, in most of the cases, the definition of a loss function that is specific for the case at hand and therefore not easily generalizable to different empirical problems. Re-balancing strategies are more general and not problem specific. That explains their great success in applied research and the focus on explaining their performances and on improving them.

Re-balancing the class sizes in the training dataset, is usually obtained either by oversampling the minority class or by under-sampling the majority class, or by a combination of both. The rebalanced data are then used to train the classifiers.

    As far as two-class linear discriminant analysis is concerned, the problem has been addressed, among others, by \cite{xie2007effect}, \cite{xue2008unbalanced}, \cite{xue2014does}.

    Through a wide simulation study supported by theoretical considerations, \cite{xue2008unbalanced} show that AUC generally favors balanced data but the increase in the median AUC for LDA after re-balancing is relatively small. On the contrary, error rate favors the original data and re-balancing causes a sharp increase in the median error rate. They also stress that re-balancing affects the performances of LDA in both the equal and unequal covariance case.

\cite{xue2014does} prove that, in the Gaussian case, using the rebalanced training data can often increase the area under the ROC curve (AUC) for the original, imbalanced test data. In particular they
demonstrate that, at least for LDA, there is an intrinsic, positive relationship
between the re-balancing of class sizes and the improvement of AUC and the largest improvement in AUC can be achieved, asymptotically, when the two classes are fully rebalanced to be of equal sizes.

	However, when the two Gaussian classes have
	similar covariance matrices, re-balancing class sizes only provides a little improvement in AUC for LDA. Moreover, re-balancing class sizes
	may not improve AUC when LDA is applied to non-Gaussian data.

	In both the above mentioned papers re-balancing is obtained either by  randomly undersampling the largest class or by randomly oversampling the smallest one.

	It has however been argued that random undersampling may  lose some relevant information while randomly oversampling with replacement the smallest class may lead to overfitting.

	To avoid these drawbacks, solutions focusing on the border between the classes have been suggested. \cite{mani2003knn} proposed selecting majority class examples whose average distance to its three nearest minority class examples is smallest. A similar approach is suggested by \cite{fithian2014local} in the context of logistic regression. They propose a method of efficient subsampling by adjusting the class balance locally in the feature space via an acceptance-rejection scheme. The proposal generalizes case-control sampling, using a pilot estimate to preferentially select examples whose responses (i.e. class membership identifiers) are conditionally rare given their features.

With special reference to classification trees and naive-Bayes classifiers Chawla \textit{et al.} (2002) propose a strategy that combines random undersampling of the majority class with a special kind of oversampling for the minority one. They try to improve upon literature results according to which undersampling the majority class leads to better classifier performance than oversampling and combining the two does not produce much improvement with respect to simple undersampling.

They design an oversampling approach which creates synthetic examples (SMOTE -  Synthetic Minority Over-sampling Technique) rather than oversampling with replacement. The minority class is over-sampled by taking each minority class sample and introducing synthetic examples along the line segments joining any/all of the $k$ minority class nearest neighbors. Depending upon the amount of over-sampling required, neighbors from the $k$ nearest neighbors are randomly chosen. 

The synthetic examples allow to create larger and less specific decision regions, thus overcoming the overfitting effect inherent in random oversampling. SMOTE oversampling is combined with majority class undersampling.

 The idea of creating synthetic examples has been followed also by \cite{menardi2014training} who proposed a method they called ROSE-\textit{Random OverSampling Examples} \citep[see, for a description of the corresponding R package,][]{lunardon2014rose}. In this solution, units from both classes are generated by resorting to a smoothed bootstrap approach. A unimodal density is centered on randomly selected observations and new artificial data are randomly generated from it.
  
  The key parameter of the procedure is the dispersion matrix of the chosen unimodal density which plays the role of smoothing parameter. The full dataset size is often kept fixed while allowing half of the units to be generated from the minority class and of half from the majority one. The method is applied to classification trees and logit models.

\section{Matrix Sketching}

Matrix sketching is a probabilistic data compression technique. Its goal is to reduce the number of rows in a data set and the task is accomplished by linearly combining the rows of the original data set through randomly generated coefficients.
The analysis can then be performed on the reduced matrix, thus saving time and space.

The theoretical justification for this approach to data compression is given by Johnson-Lindenstrauss lemma \citep{johnson1984extensions}.

\begin{lemma}
	\textbf{Johnson Lindenstrauss (1984)}.
	Let $Q$ be a subset of $p$ points in $\mathbb{R}^{n}$, then for
	any $\epsilon \in (0,1/2)$  and for $k=\dfrac{20 \, log p}{\epsilon^{2}}$ there exists a Lipschitz mapping $f:\mathbb{R}^{n} \longrightarrow \mathbb{R}^{k}$ such that for all $\mathbf{u}$, $\mathbf{v}$ $\in$ $Q$:
	$$(1-\epsilon) \rVert \mathbf{u}-\mathbf{v} \rVert^{2} \leq \rVert f(\mathbf{u})-f(\mathbf{v}) \rVert^{2} \leq (1+\epsilon) \rVert \mathbf{u}-\mathbf{v} \rVert^{2} $$
\end{lemma}
The Lemma says that any $p-$point subset of the Euclidean space can be embedded in $k$ dimensions without distorting the
distances between any pair of points by more than a factor of $ 1\pm \epsilon$, for any $\epsilon$ in $(0,1/2)$.\\
Moreover, it also gives an explicit bound on the dimensionality required for a projection to ensure that it will approximately preserve distances. This bound depends on the dimension of the data matrix that is not sketched, i.e. $p$ in this case.
\\The original proof by Johnson and Lindenstrauss is probabilistic, showing that projecting the $p$-point subset onto a random $k$-dimensional subspace only changes the inter point distances by $ 1\pm \epsilon$ with positive probability.

In order to apply Johnson and Lindenstrauss lemma, the concept of $\epsilon-$subspace embedding is useful.
\begin{definition}
	$\mathbf{\epsilon-}$\textbf{subspace embedding.}
	For a given $n \times p$ matrix $\mathbf{X}$, we call a $k\times n$ random matrix $\mathbf{S}$
	an $\epsilon$-subspace for $\mathbf{X}$, if for all vectors $\mathbf{z} \in \mathbb{R}^{p}$
	$$(1-\epsilon) \rVert \mathbf{X} \, \mathbf{z} \rVert^{2} \leq \rVert \mathbf{S} \,\mathbf{X} \, \mathbf{z} \rVert^{2} \leq (1+\epsilon) \rVert \mathbf{X} \, \mathbf{z} \rVert^{2} $$
\end{definition}
$\mathbf{S}$ is usually called \textit{Sketching Matrix}. It reduces the sample size from $n$ to $k$ whilst preserving much of the linear information in
the full dataset. As a consequence of Johnson and Lindenstrauss Lemma the scalar product too is preserved after random projections.

The original proof by Johnson and Lindenstrauss required $\mathbf{S}$ to have orthogonal rows; subsequent proofs relaxed the orthogonality requirement and assumed the entries of $\mathbf{S}$ to be independently randomly generated from a Gaussian distribution with $0$ mean and variance equal to $1/k$.
This approach to sketching is known as Gaussian sketching and it is largely used in statistical applications as it allows for statistical analysis of the results obtained after sketching.

Although appealing from a theoretical point of view, Gaussian sketching is computationally demanding as the associated sketching matrix is full. Therefore research has been oriented towards developing more efficient algorithms still satisfying the $\epsilon$-subspace embedding property.

\cite{ailon2009fast} have proposed what is known as Hadamard sketch. The sketching
matrix is formed as $\mathbf{S} = \Phi \mathbf{H D}/\sqrt{k}$ where $\Phi$ is a $k \times n$ matrix
and $\mathbf{H}$ and $\mathbf{D}$ are both $n \times n$ matrices.
The fixed matrix $\mathbf{H}$ is a Hadamard matrix of order $n$. A Hadamard matrix is a square matrix
with elements that are either $+1$ or $-
1$ and orthogonal rows. Hadamard matrices do not exist
for all integers $n$, the source dataset can be padded with zeros so that a conformable Hadamard
matrix is available. The random matrix $\mathbf{D}$ is a diagonal matrix where each nonzero element is
an independent Rademacher random variable. The random matrix $\Phi$ subsamples $k$ rows of $\mathbf{H}$
with replacement. The structure of the Hadamard sketch allows for fast matrix multiplication,
reducing calculation of the sketched dataset from $O(npk)$ of the gaussian sketch to $O(np \log k)$ operations.

Another efficient method for generating $\epsilon-$subspace embeddings is the so-called Clarkson-Woodruff sketch \citep{clarkson2017low}. The sketching matrix is a sparse random matrix $\mathbf{S=} \Gamma \, \mathbf{D}$,
where $\Gamma (k \times n) $ and $\mathbf{D} (n \times n)$ are two independent random matrices.
The matrix $\Gamma$ is a random matrix
with only one element for each column set to +1. The matrix \textbf{D} is the
same as above.
This results in a sparse random matrix $\mathbf{S}$ with only one nonzero entry per column. The sparsity speeds up matrix multiplication, dropping the complexity of generating the sketched dataset to $O(np)$.

It is worth noticing that the rows of the Gaussian and Clarkson-Woodruff sketching matrices are not orthogonal and this implies that the geometry of the original space is not preserved after sketching. The Gaussian sketching matrix is sometimes orthogonalized according to Gram-Schmidt procedure thus leading to what are known as Haar projections. This operation inevitably increases the computational load. Hadamard sketching matrices on the contrary are orthogonal by construction.

Sketching methods have mainly been used as a data compression technique in the context of multiple linear regression, where the computation of the Gram matrix may become especially demanding for large $n$ \citep{ahfock2017statistical,woodruff2014sketching}. In \cite{falcone2019} the use of the method has been extended to supervised classification.

\section{Rebalancing through Sketching}
Linear Discriminant Analysis (LDA) can be seen as a particular regression problem, therefore the same computational load inherent in the computation of the Gram matrix involved in multiple linear regression also affects LDA.

Given two groups $\mathcal{G}_0$ and $\mathcal{G}_1$, of size $n_0$ and $n_1$, coming from two omoschedastic populations with common covariance matrix $\mathbf{\Sigma}$, let $\mathbf{X}_0$ and $\mathbf{X}_1$ denote the mean centered data matrices of population null and one, respectively, and $\bar{\textbf{x}}_0$, $\bar{\textbf{x}}_1$ the corresponding mean vectors, where the subscript 1 identifies the minority class.

It is well known \citep{fisher1936use,anderson1962introduction,mclachlan2004discriminant}, the only linear discriminant direction is:
\[ \mathbf{a}= \mathbf{W}^{-1}(\bar{\mathbf{x}}_1-\bar{\mathbf{x}}_0), \]

where $\mathbf{W}$, the within group covariance matrix, is

\begin{equation}\label{eq:within}
\mathbf{W}=(\mathbf{X}_0^{\top}\mathbf{X}_0+\mathbf{X}_1^{\top}\mathbf{X}_1)/(n_0+n_1-2).
\end{equation}

After sketching the two groups separately, the sketched discriminant direction can be written as:

\begin{equation}
\label{eq:a_p}
    \mathbf{a}_{sk}= \mathbf{W}_{sk}^{-1} (\bar{\mathbf{x}}_1-\bar{\mathbf{x}}_0)=
    (n_0+n_1-2)(\tilde{\mathbf{X}}_0^{\top}\tilde{\mathbf{X}}_0+\tilde{\mathbf{X}}_1^{\top}\tilde{\mathbf{X}}_1)^{-1}(\bar{\mathbf{x}}_1-\bar{\mathbf{x}}_0)
    \end{equation}

Sketching preserves the scalar product while reducing the data set size. As the sketched data are obtained through random linear combinations of the original ones, most of the linear information is preserved after sketching. This means that, in the imbalanced data case, the size of the majority class can be reduced through sketching without incurring the risk of losing (too much) linear information.

Sketching the majority class can therefore be considered as a theoretically sound alternative to majority class undersampling. The sketched majority class data matrix reduced to the size of the minority class will then be
$\widetilde{X}_{0}$ ($n_1 \times p$) and the corresponding covariance matrix will be $\mbox{Var}(\widetilde{X}_{0})=  (\widetilde{X}_{0}^{\top}  \widetilde{X}_{0})/(n_{0}-1)$.\\


Sketching has been proposed as a data compression technique but, as a consequence of Johnson Lindenstrauss lemma, the scalar product preservation also holds when the sketching matrix has a number of rows that is larger than the number of original data points. This allows to think of this unconventional way of using sketching as an alternative to random oversampling that generates synthetic new examples from the minority class (through random linear combination of all of them) while preserving the linear structure in the data. This allows to enlarge the decision area and thus to avoid overfitting. For example, in case it is desired to increase the size of the minority class so that it equals the one of the majority class the ``oversketched" minority class data matrix will be $\widetilde{X}_{1}$ ($n_0 \times p$) and the corresponding covariance matrix will therefore be: $\mbox{Var}(\widetilde{X}_{1	})= (\widetilde{X}_{1}^{\top}  \widetilde{X}_{1})/(n_{1}-1)$ \\
The rebalanced covariance matrices can then be plugged into the within group covariance matrix and used for the computation of a new linear discriminant direction.\\
Sketching the majority class and oversketching the minority one can also be used in a combined way.\\

The use of matrix sketching is undoubtedly coherent with linear discriminant analysis which is based on the Gram matrix. Its performance in combination with other classification methods is not supported by the same strong theoretical motivation and can only be assessed through empirical analysis. This will be the topic of the next section in which different sketching methods are compared.

All the different sketching methods preserve the Gram matrix even if with a different goodness of approximation for different degrees of sketching. The different sketching methods can however change the data distribution. For instance Gaussian sketching tends to ``gaussianize" the data and can therefore strongly distort skew data distributions. Moreover, as each linear combination is a function of all the units, potentially outlying observations impact on all the sketched data values and their effect is amplified. This effect is less evident for instance for Clarkson-Woodruff sketching which, being a sparse sketching method, only selects a few units for each random linear combination.

\section{Empirical Results}
The properties of sketching as a re-balancing method have been tested on many real datasets which differ in terms of imbalance degree. Here we report the results on the two most significant ones (\texttt{spine} and \texttt{mammography}) which have been classified both by linear discriminant analysis (see Tables \ref{table:1} and \ref{table:2}) and C4.5 classification tree \citep{quinlan2014c4} (see Tables \ref{table:3} and \ref{table:4}). The data set \texttt{mammography}
\citep{woods1993} has 6 attributes and 11,183 samples that are labeled as noncalcification and calcifications (available at \url{https://www.openml.org/d/310}). The data set \texttt{spine} \citep{Dua2019} is composed of $p=6$ biomechanical features used to classify $n=310$ orthopedic patients into 2 classes, normal or abnormal (available at \url{http://archive.ics.uci.edu/ml/datasets/vertebral+column}).

Gaussian, Hadamard and Clarkson-Woodruff sketching have been applied in order to reduce the size of the majority class to the one of the minority class and in order to increase the size of the minority class so that it is as large as the majority class one. They have also been jointly used so that the size of both classes is twice the minority class size. For this last case re-balancing through SMOTE is also performed. For comparison, ROSE with its default option of preserving the total size is considered too.

Each data set has been randomly split in two parts: $75\%$ of the units for both classes constituted the training set and the remaining $25\%$ formed the test set. The procedure was repeated 200 times. The values in the tables represent the median of the quantity of interest over the 200 replicates.

The performance of the classifiers has been measured in terms of accuracy, sensitivity, specificity and area under the ROC curve (AUC).

The code implementing our procedure is available on request; ROSE and SMOTE have been applied using the corresponding R packages
\texttt{ROSE} and \texttt{DMwR}.
			\begin{table}[!h]
				\centering
					\caption{\texttt{{spine}} dataset, $n$=310, $\pi_1$=32\%  - Median values (over 200 replications)}
				\label{table:1}
				\medskip
			\begin{tabular}{lrrrr}
				& Accuracy & Sensitivity & Specificity & AUC \\
				\toprule
				LDA & 0.831 & 0.904 & 0.680 & 0.897 \\
				\midrule
				Under-Sampling     & 0.792 & 0.731& 0.920& 0.907 \\
				Gauss Partial Sk   & 0.792 & 0.731& 0.920& 0.900 \\
				CW Partial Sk      & 0.792& 0.750& 0.880& 0.899 \\
				Hada Partial Sk    & 0.792 & 0.731& 0.920& $\mathbf{0.910}$ \\
				\midrule
				Over-Sampling      & 0.792 & 0.750& 0.920& 0.900 \\
				Gauss Partial OverSk  & 0.792 & 0.731& 0.920& 0.903 \\
				CW    Partial OverSk  & 0.792& 0.731& 0.920& 0.902 \\
				Hada  Partial OverSk  & 0.792 & 0.731& 0.920& $\mathbf{0.910}$ \\
				\midrule
				ROSE& 0.792 & 0.712 & 0.960 & 0.905 \\
				\midrule
				SMOTE      & 0.792 & 0.750& 0.920& 0.905 \\
				UndOver-Sampling Bal& 0.792 & 0.750& 0.900& 0.902 \\
				Gauss Bal Sk & 0.792 & 0.750& 0.880& 0.896 \\
				CW Bal Sk    & 0.792& 0.750& 0.920& 0.902 \\
				Hada Bal Sk  & 0.792 & 0.731& 0.920& $\mathbf{0.910}$ \\
				\bottomrule
			\end{tabular}
			\end{table}

		\begin{table}[!h]
			\centering
			\caption{\texttt{{mammography}},  $n$=11,183, $\pi_1$=2.32\%  - Median values (over 200 replications)}
				\label{table:2}
				\medskip
			\begin{tabular}{lrrrr}
				& Accuracy & Sensitivity & Specificity & AUC \\
				\toprule
				LDA    & 0.977& 0.986& 0.554& 0.903\\
				\midrule
				Under-Sampling     & 0.830& 0.829& 0.892& 0.928\\
				Gauss Partial Sk   & 0.827& 0.826& 0.907& $\mathbf{0.930}$\\
				CW Partial Sk      & 0.827& 0.825& 0.892& 0.923\\
				Hada Partial Sk    & 0.742& 0.739& 0.892& 0.914\\
				\midrule
				Over-Sampling      & 0.829& 0.828& 0.892& 0.931\\
				Gauss Partial Over Sk  & 0.828& 0.826& 0.892& $\mathbf{0.932}$\\
				CW    Partial Over Sk  & 0.828& 0.826& 0.900& 0.931\\
				Hada  Partial Over Sk  & 0.743& 0.739& 0.892& 0.914\\
				\midrule
				ROSE       & 0.977& 1.000& 0.000& 0.878\\
				\midrule
				SMOTE      & 0.841& 0.840& 0.892& 0.930\\
				UndOver-Sampling Bal& 0.837& 0.836& 0.892& 0.928\\
				Gauss Bal Sk  & 0.829& 0.827& 0.900& $\mathbf{0.932}$\\
				CW Bal Sk     & 0.957& 0.964& 0.677& 0.900\\
				Hada Bal Sk  & 0.744& 0.740& 0.892& 0.914\\
				\bottomrule
			\end{tabular}
		\end{table}
\newpage
Table \ref{table:1} and \ref{table:2} show that, coherently with the findings in Xue and Titterington and Xue and Hall, when combined with LDA, rebalancing causes a strong decrease in the accuracy which is combined with a little increase in the AUC. However a strong increase in specificity, i.e. in the ability to correctly identify the minority class, is worth of note. In this context sketching based methods always outperform the other rebalancing methods. It does not seem to be any evidence of a systematic predominance of over, under or balanced sketching strategies.

As already said, sketching preserves the linear structure which is the core element of LDA. The good performances of sketching in this context are therefore coherent with its theoretical properties. However, when sketching methods are combined with classification methods that do not rely on the linear structure in the data, results are not so clear-cut and they seem to be strongly related to specific characteristics of the data.

Tables \ref{table:3} and \ref{table:4} report the results of C4.5 classification trees. While for the \texttt{spine} dataset the sketching methods perform well, for the \texttt{mammography} dataset sketching methods are strongly outperformed by standard random oversampling or undersampling methods and by ROSE.

\begin{table}[!h]
		\centering
		\caption{\texttt{{ spine}} dataset, $n$=310, $\pi_1$=32\%  - Median values (over 200 replications)}
		\label{table:3}
		\medskip
			\begin{tabular}{lrrrr}
				& Accuracy & Sensitivity & Specificity & AUC \\
				\toprule
				C4.5 Tree & 0.818 & 0.904 & 0.680 & 0.773 \\
				\midrule
				Under-Sampling     & 0.805 & 0.788& 0.840& 0.822 \\
				Gauss Partial Sk   & 0.792 & 0.731& 0.920& $\mathbf{0.825}$ \\
				CW Partial Sk      & 0.792& 0.750& 0.880& 0.817 \\
				Hada Partial Sk    & 0.805 & 0.750& 0.920& $\mathbf{0.825}$ \\
				\midrule
				Over-Sampling      & 0.818 & 0.846& 0.720& 0.795 \\
				Gauss Partial OverSk  & 0.805 & 0.788& 0.840& 0.813 \\
				CW    Partial OverSk  & 0.805& 0.788& 0.840& 0.815 \\
				Hada  Partial OverSk  & 0.805 & 0.769& 0.880& $\mathbf{0.825}$ \\
				\midrule
				ROSE& 0.792 & 0.712 & 0.960 & $\mathbf{0.835}$ \\
				\midrule
				SMOTE      & 0.805 & 0.808 & 0.800& 0.805 \\
				UndOver-Sampling Bal& 0.812 & 0.846& 0.760& 0.794 \\
				Gauss Bal Sk & 0.805 & 0.750& 0.920& $\mathbf{0.835}$ \\
				CW Bal Sk    & 0.805& 0.788& 0.880& 0.824 \\
				Hada Bal Sk  & 0.818 & 0.779& 0.880& 0.834 \\
				\bottomrule
			\end{tabular}
		\end{table}

	\begin{table}[!h]
\centering
\caption{\texttt{{ mammography}},  $n$=11,183, $\pi_1$=2.32\%  - Median values (over 200 replications)}
\label{table:4}
\medskip
			\begin{tabular}{lrrrr}
				& Accuracy & Sensitivity & Specificity & AUC \\
				\toprule
				C4.5 Tree & 0.985 & 0.997 & 0.508 & 0.752 \\
				\midrule
				Under-Sampling     & 0.893 & 0.894& 0.846& $\mathbf{0.879}$ \\
				Gauss Partial Sk   & 0.763 & 0.241& 0.938& 0.592 \\
				CW Partial Sk      & 0.838 & 0.165& 0.969& 0.562 \\
				Hada Partial Sk    & 0.807 & 0.806& 0.846& 0.827 \\
				\midrule
				Over-Sampling      & 0.979 & 0.987& 0.646& $\mathbf{0.815}$ \\
				Gauss Partial OverSk  & 0.964 & 0.973& 0.477& 0.729 \\
				CW    Partial OverSk  & 0.977& 0.988& 0.538& 0.762 \\
				Hada  Partial OverSk  & 0.931 & 0.937& 0.692& 0.813 \\
				\midrule
				ROSE& 0.901& 0.903 & 0.800 & 0.850 \\
				\midrule
				SMOTE      & 0.914 & 0.916 & 0.846& $\mathbf{0.879}$ \\
				UndOver-Sampling Bal& 0.914 & 0.917& 0.831& 0.873 \\
				Gauss Bal Sk & 0.720 & 0.624& 0.908& 0.724 \\
				CW Bal Sk    & 0.756& 0.257& 0.923& 0.591 \\
				Hada Bal Sk  & 0.839 & 0.838& 0.815& 0.830 \\		
				\bottomrule
			\end{tabular}
		\end{table}
\clearpage
\bibliographystyle{chicago}
\bibliography{Bibliography}

\begin{thebibliography}{}

\bibitem[\protect\citeauthoryear{Ahfock, Astle, and Richardson}{Ahfock
  et~al.}{2017}]{ahfock2017statistical}
Ahfock, D., W.~J. Astle, and S.~Richardson (2017).
\newblock Statistical properties of sketching algorithms.
\newblock {\em arXiv preprint arXiv:1706.03665\/}.

\bibitem[\protect\citeauthoryear{Ailon and Chazelle}{Ailon and
  Chazelle}{2009}]{ailon2009fast}
Ailon, N. and B.~Chazelle (2009).
\newblock The fast johnson--lindenstrauss transform and approximate nearest
  neighbors.
\newblock {\em SIAM Journal on computing\/}~{\em 39\/}(1), 302--322.

\bibitem[\protect\citeauthoryear{Anderson}{Anderson}{1962}]{anderson1962introduction}
Anderson, T.~W. (1962).
\newblock {\em An introduction to multivariate statistical analysis}.
\newblock Wiley New York.

\bibitem[\protect\citeauthoryear{Chawla, Japkowicz, and Kotcz}{Chawla
  et~al.}{2004}]{chawla2004special}
Chawla, N.~V., N.~Japkowicz, and A.~Kotcz (2004).
\newblock Special issue on learning from imbalanced data sets.
\newblock {\em ACM Sigkdd Explorations Newsletter\/}~{\em 6\/}(1), 1--6.

\bibitem[\protect\citeauthoryear{Clarkson and Woodruff}{Clarkson and
  Woodruff}{2017}]{clarkson2017low}
Clarkson, K.~L. and D.~P. Woodruff (2017).
\newblock Low-rank approximation and regression in input sparsity time.
\newblock {\em Journal of the ACM (JACM)\/}~{\em 63\/}(6), 54.

\bibitem[\protect\citeauthoryear{Dua and Graff}{Dua and Graff}{2019}]{Dua2019}
Dua, D. and C.~Graff (2019).
\newblock {UCI} machine learning repository.

\bibitem[\protect\citeauthoryear{Falcone}{Falcone}{2019}]{falcone2019}
Falcone, R. (2019).
\newblock {\em Supervised Classification with Matrix Sketching}.
\newblock Ph.\ D. thesis, University of Bologna.

\bibitem[\protect\citeauthoryear{Fisher}{Fisher}{1936}]{fisher1936use}
Fisher, R.~A. (1936).
\newblock The use of multiple measurements in taxonomic problems.
\newblock {\em Annals of eugenics\/}~{\em 7\/}(2), 179--188.

\bibitem[\protect\citeauthoryear{Fithian and Hastie}{Fithian and
  Hastie}{2014}]{fithian2014local}
Fithian, W. and T.~Hastie (2014).
\newblock Local case-control sampling: Efficient subsampling in imbalanced data
  sets.
\newblock {\em Annals of statistics\/}~{\em 42\/}(5), 1693.

\bibitem[\protect\citeauthoryear{Johnson and Lindenstrauss}{Johnson and
  Lindenstrauss}{1984}]{johnson1984extensions}
Johnson, W.~B. and J.~Lindenstrauss (1984).
\newblock Extensions of lipschitz mappings into a hilbert space.
\newblock {\em Contemporary mathematics\/}~{\em 26\/}(189-206), 1.

\bibitem[\protect\citeauthoryear{Lunardon, Menardi, and Torelli}{Lunardon
  et~al.}{2014}]{lunardon2014rose}
Lunardon, N., G.~Menardi, and N.~Torelli (2014).
\newblock Rose: A package for binary imbalanced learning.
\newblock {\em R journal\/}~{\em 6\/}(1).

\bibitem[\protect\citeauthoryear{Mani and Zhang}{Mani and
  Zhang}{2003}]{mani2003knn}
Mani, I. and I.~Zhang (2003).
\newblock knn approach to unbalanced data distributions: a case study involving
  information extraction.
\newblock In {\em Proceedings of workshop on learning from imbalanced
  datasets}, Volume 126.

\bibitem[\protect\citeauthoryear{McLachlan}{McLachlan}{2004}]{mclachlan2004discriminant}
McLachlan, G. (2004).
\newblock {\em Discriminant analysis and statistical pattern recognition},
  Volume 544.
\newblock John Wiley \& Sons.

\bibitem[\protect\citeauthoryear{Menardi and Torelli}{Menardi and
  Torelli}{2014}]{menardi2014training}
Menardi, G. and N.~Torelli (2014).
\newblock Training and assessing classification rules with imbalanced data.
\newblock {\em Data Mining and Knowledge Discovery\/}~{\em 28\/}(1), 92--122.

\bibitem[\protect\citeauthoryear{Quinlan}{Quinlan}{2014}]{quinlan2014c4}
Quinlan, J.~R. (2014).
\newblock {\em C4. 5: programs for machine learning}.
\newblock Elsevier.

\bibitem[\protect\citeauthoryear{Woodruff}{Woodruff}{2014}]{woodruff2014sketching}
Woodruff, D.~P. (2014).
\newblock Sketching as a tool for numerical linear algebra.
\newblock {\em Foundations and Trends{\textregistered} in Theoretical Computer
  Science\/}~{\em 10\/}(1--2), 1--157.

\bibitem[\protect\citeauthoryear{Woods, Doss, Bowyer, Solka, Priebe, and
  Kegelmeyer}{Woods et~al.}{1993}]{woods1993}
Woods, K.~S., C.~C. Doss, K.~W. Bowyer, J.~L. Solka, C.~E. Priebe, and W.~P.
  Kegelmeyer (1993).
\newblock comparative evaluation of pattern recognition techniques for
  detection of microcalcifications in mammography.
\newblock {\em International Journal of Pattern Recognition and Artificial
  Intelligence\/}~{\em 07\/}(06), 1417--1436.

\bibitem[\protect\citeauthoryear{Xie and Qiu}{Xie and
  Qiu}{2007}]{xie2007effect}
Xie, J. and Z.~Qiu (2007).
\newblock The effect of imbalanced data sets on lda: A theoretical and
  empirical analysis.
\newblock {\em Pattern recognition\/}~{\em 40\/}(2), 557--562.

\bibitem[\protect\citeauthoryear{Xue and Hall}{Xue and
  Hall}{2014}]{xue2014does}
Xue, J.-H. and P.~Hall (2014).
\newblock Why does rebalancing class-unbalanced data improve auc for linear
  discriminant analysis?
\newblock {\em IEEE transactions on pattern analysis and machine
  intelligence\/}~{\em 37\/}(5), 1109--1112.

\bibitem[\protect\citeauthoryear{Xue and Titterington}{Xue and
  Titterington}{2008}]{xue2008unbalanced}
Xue, J.-H. and D.~M. Titterington (2008).
\newblock Do unbalanced data have a negative effect on lda?
\newblock {\em Pattern Recognition\/}~{\em 41\/}(5), 1558--1571.

\end{thebibliography}

\end{document}